\documentclass{ifacconf}
\usepackage{tabularx, multirow, booktabs}
\usepackage{graphicx}     
\usepackage{natbib}        
\begin{document}
\begin{frontmatter}

\title{ A Survey on Data-Driven Fault Diagnostic Techniques for Marine Diesel Engines} 

\author[First]{A. Youssef}
\author[First]{H. Noura}
\author[First]{A. EL Amrani}
\author[First]{E. El Adel}
\author[First]{M. Ouladsine}
\address[First]{Aix Marseille Univ., CNRS, LIS (UMR 7020), Avenue Escadrille Normandie-Niemen, F-13397 Marseille Cedex 20, France (e-mail : ayah.youssef@lis-lab.fr, hassan.noura@lis-lab.fr, abderrahim.el-amrani@lis-lab.fr, el-mostafa.el-adel@lis-lab.fr, mustapha.ouladsine@lis-lab.fr).}

\begin{abstract}                

Fault diagnosis in marine diesel engines is vital for maritime safety and operational efficiency. These engines are integral to marine vessels, and their reliable performance is crucial for safe navigation. Swift identification and resolution of faults are essential to prevent breakdowns, enhance safety, and reduce the risk of catastrophic failures at sea. Proactive fault diagnosis facilitates timely maintenance, minimizes downtime, and ensures the overall reliability and longevity of marine diesel engines. This paper explores the importance of fault diagnosis, emphasizing subsystems, common faults, and recent advancements in data-driven approaches for effective marine diesel engine maintenance.
\end{abstract}


\end{frontmatter}

\section{Introduction}
\vspace{-0.4cm}
After the COVID-19 pandemic and the Ukraine War, maritime trade volume contracted marginally by 0.4\% in 2022. However, according to projections from the United Nations Conference on Trade and Development (UNCTAD), it is expected to grow by 2.4\% in 2023. Additionally, in 2023, oil cargo distances reached long-term highs due to disruptions caused by the Russian Federation's search for new export markets for crude oil and refined products, while Europe sought alternative energy suppliers (\cite{unctad2023}). These developments have led to an increase in operational hours and cargo transportation for merchants, exposing vessels to prolonged periods of high loads, high humidity, and vibrational environments, which in turn increase the risk of malfunctions on ships. Of all the systems on a vessel platform, the main engine plays the most critical role as it is responsible for propulsion and power. Most vessels rely on diesel engines, which are renowned for their fuel efficiency (\cite{tadros2020}). In 2022, the marine diesel engine market surpassed a valuation of USD 4.7 billion. It is expected to experience a Compound Annual Growth Rate (CAGR) of 4.8\% from 2023 to 2032. This growth is primarily driven by advancements in emerging economies and the continual expansion of seaborne trade (\cite{gmi2023}). The statistics of the damage of the ship equipment indicate that marine diesel engines are more vulnerable to damage than other machines and devices (\cite{witkowski2020}). The significance of maintaining the main engine in optimal condition cannot be overstated, as any fault in this critical component can lead to severe consequences, potentially even collisions. Therefore, the development of fault diagnostic techniques is imperative to detect and address any issues before they escalate and cause damage to the engine or other components and ensure safety and reliability during the operation process (\cite{jiang2017}, \cite{szymanski2016}). The evolution of diagnostic techniques for marine diesel engines can be categorized into three distinct stages spanning from the 1960s to the present day. The first stage, spanning from the 1960s to the 1990s, involved field testing techniques, which were traditional methods that required acquired signals to undergo further processing to extract relevant fault features. The second stage, from the 1990s to 2010, introduced online condition monitoring and remote fault diagnosis, allowing for real-time fault detection through monitoring and alarm systems, which helped identify abnormal states between periodic checks. The most recent stage, from 2010 to the present, marks the era of Intelligent Fault Diagnosis. This stage incorporates advanced automation, artificial intelligence (AI), and big data analysis, with diagnostic techniques relying on quantitative models or data-driven approaches, as well as expert systems. It is important to note that the complexity of the marine diesel engine system entails that not all faults can be effectively diagnosed using the same technique. Therefore, a tailored approach is necessary to address the diverse range of potential malfunctions in this critical component. 
Advanced fault diagnostic techniques not only allow for the early identification of issues but also facilitate data-driven insights and predictive maintenance strategies. They enable ship operators to minimize downtime, lower maintenance costs, and reduce the environmental footprint of their vessels, aligning with international regulations and sustainability goals. This paper discusses and reviews the recent data-based techniques used for fault diagnosis on marine diesel engines.
In Section \ref{sec:DE}, a brief introduction to marine diesel engines is provided through a description of engine subsystems functioning. Section \ref{sec:Fault Diagnosis Techniques} is devoted to the analysis of different diagnostic techniques used for the diagnosis of subsystems. Section \ref{sec: Data driven methods} analyzes and lists recent data-driven methods used for fault diagnosis on different marine diesel engine subsystems. Finally, concluding remarks are given in Section \ref{sec:conclusion}.

\section{Description of Diesel Engine} \label{sec:DE}
\subsection{Marine Diesel Engine}
\begin{figure}
    \centering
    \includegraphics[width=7.8cm, height=5cm]{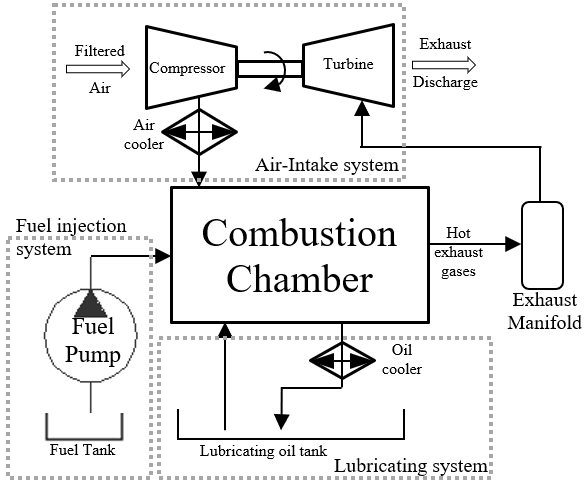} 
    \caption{Marine Diesel Subsystems}
    \label{fig:Subsystems1}
\end{figure}

\vspace{-0.4cm}
The diesel engine, credited to Rudolph Diesel's invention in 1897, stands as a pivotal internal combustion engine, serving as the powerhouse of modern industry by converting fuel into energy. In 1903 the first marine diesel engine was built (\cite{paul2020}). Marine diesel engines are large in scale and more complex than those used in automobiles and safety issues they are more critical where they are subjected to a higher risk of failure due to the surrounding environment, and it is difficult to repair and maintain diesel engines during maritime shipping. Mechanical failure may cause the diesel engine to stop the supply of ship power and even the unit to be scrapped (\cite{kamaltdinov2019experimental}, \cite{huo2020adaptive}). These engines operate on the principle of compression ignition, where the fuel ignites due to the high temperature and pressure generated during air compression within the engine's cylinders. The primary subsystems of a diesel engine, as depicted in Fig. \ref{fig:Subsystems1}, are the fuel injection system, cooling system, lubrication system, and air intake and exhaust system (\cite{cai2017}). These subsystems collaborate seamlessly to ensure the efficient and reliable operation of marine diesel engines. 
The air intake system is entrusted with the critical task of supplying and filtering the air needed for combustion, while the fuel injection system precisely controls the delivery of fuel into the engine's cylinders where dual diaphragm design is used to create a fail-safe if that compartment fails while for industrial engines single diaphragm diesel pump is used. Concurrently, the cooling system plays a pivotal role in maintaining the engine's temperature within safe operating limits to prevent overheating, water and air cooling is used for cooling down the engine. Furthermore, the exhaust system takes on the responsibility of effectively eliminating combustion by-products and managing emissions in compliance with environmental standards, the lubricating system is responsible for reducing friction between moving components and cleaning engine parts by removing contaminants and dampening noise and vibration.

\subsection{Faults on Marine Diesel Engines}
\vspace{-0.4cm}
Faults can occur in various components of the engine and are typically classified based on their type: Control System Faults (including Hardware and Software faults) or Actuator Faults (such as Mechanical or Hydraulic faults). It is important to recognize that the interdependence of subsystems within the engine means that a fault originating in one part may have a cascading effect, impacting other parts. Identifying the cause-and-effect link between system faults and their observable symptoms serves as the initial phase in the process of detecting and isolating faults (\cite{luo2021dynamic}, \cite{singh2018analytical}).
\section{Fault Diagnostic Techniques}\label{sec:Fault Diagnosis Techniques}
\vspace{-0.4cm}
The United States was the pioneering country in developing fault diagnosis technology. In 1967, NASA and the Naval Research Institute jointly established the first mechanical prevention team. The ability to predict malfunctions before they occur allows for cost reduction by transitioning from planned maintenance to predictive maintenance. Diagnostic techniques in this field can be broadly categorized into three groups: model-based, qualitative empirical knowledge-based, and data-based methods (\cite{qin2012}). 
\begin{enumerate}
\item Model-based techniques: diagnose the working status by analyzing residuals, using parity space, observer-based approach and parameter estimation approach. \cite{Fu2023} developed a physical model for a marine diesel engine for diagnosis by calculating the residuals between predicted data variables using their models and measured data. However, due to the complexity of marine engine systems, establishing an accurate mathematical model to maintain robustness and sensitivity to faults has limitations, especially in dealing with complex systems.
\item Empirical knowledge-based technique: exemplified by Expert Systems, heavily relies on the specialized expertise of engineers for diagnostic purposes. Various types of expert systems, such as rule-based, case-based, framework, fuzzy logic, neural networks, and genetic algorithms, are employed for diagnosis (\cite{Sahin2012}). Fuzzy expert systems, in particular, are widely utilized in this domain as developed by \cite{tasdemir2011}. However, this method exhibits limited efficacy in detecting novel or unexpected problems, necessitating continual modifications and development of the knowledge database.
\item Data-based technique: employs data to build models, enhancing the performance of diesel engine fault monitoring systems through training experiences. This method relies on abundant measured data where no prior knowledge about the investigated object is required. The data-driven learning techniques can be categorized into two paradigms: supervised learning and unsupervised learning.
\end{enumerate}
Advanced diagnostic techniques enable the prediction of malfunctions before they occur, reducing costs by transitioning from planned maintenance to predictive maintenance (\cite{kocak2023}). Various data-driven diagnostic techniques, including Support Vector Machine (SVM), Principle Component Analysis (PCA), Non-Negative Matrix Factorization (NMF), Neural Networks (NN) based on deep learning, and others, are employed for the diagnosis of marine diesel engines.
The data-driven diagnostic technique typically involves three stages: pre-processing, domain-adaptive training, and fault diagnosis. Pre-processing of data takes up to 50\% to 80\% of the whole process where data is treated depending on the type of data found by data transformation, information gathering or gathering new information (\cite{Maharana2022}). In training and validation, data is trained to build the model and then tested, and finally, the algorithm is ready for diagnosis.

\section{Data-Driven Diagnosis on Marine Diesel Engines} \label{sec: Data driven methods}
\vspace{-0.4cm}
Researchers are increasingly turning their attention to utilizing sensor data, operational parameters, and ML for improved detection and prediction of issues in marine diesel engines. This approach allows for proactive and accurate identification of faults in these engines. Given the intricacies of marine diesel engines and the vast array of sensors capturing data from various subsystems, each subsystem is diagnosed individually due to the diverse conditions of data within them.
\subsection{Fuel Injection System}
\vspace{-0.4cm}
The most critical component of a diesel engine is its fuel injection equipment. Even minor faults in this system can result in a significant loss of combustion efficiency, increased engine emissions, and elevated noise levels (\cite{krogerus2016}). The fuel injection system serves the crucial role of transferring fuel from the fuel tank, typically via a fuel pump, to the fuel injectors responsible for delivering the precise amount of fuel required for combustion within each cylinder. When considering damage within the injection system, the majority of issues are related to the following components: injectors, accounting for 41\% of reported problems; injector pumps, responsible for 31\% of issues; and fuel pipes, which contribute to 12\% of identified faults (\cite{witkowski2020}). Monitoring such faults is essential to protect the engine,  ensure better performance and reduce pollutants. Many research studies have been conducted to diagnose various issues occurring in the injection system. \cite{shi2023a} used SVM method to diagnose injection faults by proposing the "Improved Refined Composite Multi-Scale Dispersion Entropy (IRCMDE)" method for feature extraction which is used to overcome the loss of useful information by Multi-scale Dispersion Entropy (MDE) and improve the accuracy of the entropy values. The IRCMDE value is computed for a time series $u$ with signal length $N$, a predefined number of classes $c$, and embedding dimension $m$. To calculate the IRCMDE value at a scale factor $\tau$, first the length of the coarse-grained time series must be found, $\frac{N}{\tau}$. This is achieved by coarse-graining the time series using an improved method, obtaining the sequence $Y^{(\tau)}_K$. Then, the Dispersion Entropy (DE) value is calculated for each sequence $y^{(\tau)}_{k,j}$ separately. Finally, the final IRCMDE value for the scale factor $\tau$ is computed using these DE values as shown in the following equation:
$IRCMDE(u,m,c,d,\tau)=\frac{1}{\tau}\sum^{\tau}_{k=1}DE({y^{(\tau)}_{k,j}},m,c,d)$

Then, for the feature selection process, the Fast Correlation-Based Filter (FCBF), a sophisticated multivariate method, is applied. FCBF utilizes a heuristic approach by applying a backward selection technique along with a sequential search strategy to systematically eliminate irrelevant and redundant features. Subsequently, the SVM method is employed for the classification of the engine's operating condition. In this approach, data undergo training and optimization through the Particle Swarm Optimization (PSO) algorithm. The classification is based on measurements derived from both fuel injection pressure and vibration signals obtained from the high-pressure pipe. To verify the method, data are collected from the MAN B\&W 6S35ME-B9 engine type, and the proposed method is compared to other methods such as RCMDE, MDE, and Multi-scale Permutation Entropy (MPE) methods, where the IRCMDE showed the highest accuracy (92.12\%) and a relatively smaller standard deviation value. The method effectively identifies issues such as delays in injection time, blocked spray holes, and worn needle valves. However, it is not conducive when classifying faults using vibration signals measured on a double-wall pipeline. \cite{hou2020} applied different PCA and optimized SVM methods and compared their accuracy by comparing the Correct Diagnosis Ratio (CDR) and Fault Misclassified Ratio (FMR). These methods are applied for fuel oil supply diagnosis where faults focused on the fuel pump. PCA with SVM, optimized by sample size, demonstrated the highest CDR of 93.9\% and the lowest FMR of 6.1\% accuracy in fault diagnosis. It effectively reduced the impact of the imbalance and high dimensionality in the fuel oil supply system, surpassing the accuracy achieved by the optimized SVM. However, this complex algorithm showed long run-time execution comparable to other methods. \cite{chen2022} used improved Genetic Algorithm Elman Neural Network (GA-ENN) adaptive fault detection technique to enhance the accuracy of the traditional GA-ENN method. The improved GA-ENN adaptive method uses a genetic algorithm to assign the weight and threshold in ENN to calculate the norm of each generation. The calculation function is given by: $f=(\sum^M_{j=1}\sum^N_{i=1}(P^2_{ji}-Z^2_{ji}))^{1/2}$

The fitness value is the error norm where $P_{ji}$ and $Z_{ji}$ are the expected and real outputs of $j$ neuron and $i$ node. This algorithm is applied so the error is to be minimal. Speed, power, maximum burst pressure, high-pressure oil pressure, exhaust pressure and temperatures of diesel fuel injection system are the input data to indicate the state of the system. The method showed 95.67\% diagnosis accuracy and showed better performance than traditional GA-ENN and SVM.

\subsection{Intake and Exhaust System}
\vspace{-0.4cm}
The intake and exhaust system of a diesel engine are crucial for its performance (\cite{abdullah2013}). In a marine diesel engine, the intake system plays a vital role in supplying clean and compressed air to the combustion chambers. Its key functions include filtering and purifying incoming air, compressing it via a turbocharger, and delivering it to the engine cylinders for mixing with fuel. This process enhances combustion efficiency, promotes engine power output, and reduces emissions, ensuring effective and reliable operation in the demanding marine environment. After combustion, the exhaust system manages the removal of exhaust gases from the cylinder part of the exhaust gas drive, powering the turbocharger of the intake system. The primary component of the exhaust system in a marine diesel engine is the exhaust valve. One common fault in the exhaust system is valve leakage, which can lead to detrimental effects, including increased exhaust temperature, reduced output power, and an overall decrease in engine performance. To diagnose faults on this subsystem without disassembling the engine, researchers have explored various methods, focusing on thermal parameters, vibration analysis, and acoustic emission (AE) signals. \cite{hu2023} conducted a study using AE signals to generalize diagnostic methods for exhaust valve leakage in marine diesel engines. The research compared supervised learning methods, specifically the principal component analysis-support vector machine (PCA-SVM) and a one-dimensional convolutional neural network (1D-CNN), with domain-adaptive networks, including deep adaptation network (DAN), domain adversarial neural network (DANN), and margin disparity discrepancy (MDD). These techniques were applied to two distinct marine diesel engines. The study investigated the diagnostic capabilities of domain-adaptive methods, emphasizing the assessment of feature transferability across different engine types. The proposed method, particularly MDD, demonstrated significant accuracy in cross-engine-type fault diagnosis, outperforming alternative methods. In harsh operating environments, vibration signals often exhibit nonstationary behavior, making it challenging to isolate fault features from substantial noise. Integrating traditional deep learning can result in significant performance losses, especially when dealing with new diagnostic tasks or limited datasets. To address these challenges, transfer learning has been proposed for exhaust valve leakage diagnosis. \cite{cai2023} introduced a modified VGG16 deep convolutional neural network transfer method, achieving an impressive 95.2 percent accuracy in diagnosing valve leakage. Another approach, presented by \cite{wang2020}, combines supervised and unsupervised learning techniques. They utilized a neural network algorithm employing PCA clustering analysis. PCA is used for the reduction of the dimensionality of the raw data, simplified through k-means clustering for grouping the sample data. The determination of the k value relies on the computation of the function $J$ defined in the following equation: $J(c,k) = \sum^m_{i=1}||x^{(i)}-k_{c^{(i)}}||^2$ aiming to simplify intricate data into a single numerical representation. This function computes the total of squared distances between each data point and the center of mass. Where, $c^{(i)}$ signifies the nearest class for the $i$-th sample among $n$ classes, strategically chosen to minimize the $J$-function. This categorization minimizes the gap between the center of mass and the sample $x^{(i)}$, continually adjusting the centroid $k_j$ to reduce $J$ for each class. The ensuing results of k-means PCA optimization are subjected to testing by a BP NN, deploying the Mean Squared Error (MSE). A smaller distance between all output vectors and the target vector results in a smaller MSE value. The training concludes when the minimum MSE is achieved. This approach has demonstrated a notable diagnostic capability, particularly in identifying faults associated with air leakage in the exhaust valve. Through optimization of the clustering analysis neural network, the diagnostic speed is enhanced, and the number of nodes and hidden layers is minimized. It's worth noting that data-driven methods often require extensive datasets comprising both normal and fault data to effectively train a fault monitoring model. However, gathering sufficient data for all types of faults can be challenging. In response, \cite{wang2021} introduced a hybrid fault monitoring technique that combines manifold learning and anomaly detection. This approach relies solely on normal condition data to train the model. Manifold learning methods, such as Multi-Dimensional Scaling (MDS) based on the concept of dissimilarity measures that quantify the discrepancy between two data points, aim to find a projection of high-dimensional data points in a lower-dimensional space that preserves pairwise distances. Locally Linear Embedding (LLE) computes a set of weights for each point that realizes a linear combination of its neighbors to evaluate the importance of the feature through a metric criterion. T-distributed Stochastic Neighbor Embedding (t-SNE) is a dimensionality reduction technique based on probability distribution to model a high-dimensional data set by a low-dimensional dataset, mapping the data points to the probability by affinities that convert similarities between data points to joint probabilities. MDS, LLE, and t-SNE are used to extract valuable information from the original dataset. These features are then fed into an improved isolation forest (iForest) anomaly detection algorithm and trained, reducing redundancy and addressing the curse of dimensionality issues. This technique was successfully applied to diagnose faults related to exhaust gas leakage and other faults and showed that the combination of t-SNE and iForest has the best performance and is better than the PCA feature selection method. It is well known, however, that the variations of the actual data are nonlinear and highly non-Gaussian, and the majority of the data cannot be described by second-order correlations. Therefore, the employment of PCA shows a very poor performance. \cite{tu2021} diagnosed faults in the air system when there is insufficient airflow by utilizing Kernel Principal Component Analysis (KPCA) for feature extraction from the data, which is a non-linear form of PCA, and employed SVM for fault classification and diagnosis. This method effectively reduced the computational cost. For turbine faults and air filter clogging, \cite{basurko2015} used Artificial Neural Networks (ANN) for the diagnosis of these faults. A feed-forward network with sigmoid as the activation function and a back-propagation method for network training were used to develop the model. The weightings of each variable are determined by sensitivity analysis to select the most sensitive for use in the ANN model. This method is applied to faults such as a clogged turbine, air cooler, and air filter/compressor, which cause an increase in fuel consumption, and showed good performance in diagnosis. For exhaust pipe blockage, \cite{zhong2019} used Semi-Supervised PCA (SSPCA) for diagnosis rather than the traditional PCA, which cannot handle the labeled data due to its unsupervised essence. SSPCA showed better performance as it makes full use of all labeled and unlabeled samples simultaneously. Fault detection occurs when exceeding the limit of the two statistics of SSPCA $T^2_{sem}$ and $Q_{sem}$ by F distribution and Kernel Density Estimation (KDE), respectively. In case of offline monitoring, and in case of online monitoring, if $T^2_{sem}$ and $Q_{sem}$ exceed their respective control limits, a fault is detected. This method showed better accuracy in detection even if pure data is fed to PCA.
\subsection{Lubricating and Cooling System}
\vspace{-0.4cm}
Ensuring the effective functioning of the cooling system is crucial as it is responsible for attaining and sustaining optimal working conditions for the engine. If the cooling system experiences damage, it can lead to a swift increase in temperature for essential engine components. This, in turn, can negatively impact lubrication, causing a loss of lubricating properties in the oil and affecting combustion. It may also trigger premature ignition of the fuel-air mixture. Ultimately, this scenario could result in excessive thermal expansion of the piston within the cylinder, often culminating in significant engine damage. However, the lubrication system's role is to ensure proper functioning, minimise wear and tear, remove heat from metal surfaces out of the engine, neutralize acids that can be extremely corrosive, clean the engine's internal surfaces from dirt, insoluble and metal particles and prevent rust and corrosion of internal surfaces. this mission is accomplished by pumping sufficient oil to components (crankshaft, camshaft, ring/cylinders, etc). Insufficient lubrication may cause wear to the components and decrease the efficiency of combustion in a cylinder, and insufficient propulsion. For this purpose, the lubrication pressure, temperature, and quantity must be maintained and any fault occurrence must be detected. \cite{wang2022integrated} used a multivariate statistics-based approach for early detection and diagnosis for the whole lubrication system with a sensor network designed to be minimal. The data set is decomposed and pre-processed according to PCA. The multivariate statistics aimed to calculate the Hotellings $T^2$ and $Q$ control parameters according to the estimated Probability Density Function (PDF) obtained by Adaptive Kernel Density Estimation (AKDE) for more realistic limit estimation. Any statistics exceeding its control limit, the system is to be degraded from the healthy condition. The abnormalities detected are the input of the Bayesian networks which will identify the root cause of the lubrication fault, this method was tested on 11 faults that may occur in the lubrication system (Lubrication pipe leakage, lubrication oil shortage and others). Monitoring all observable parameters in large mechanical systems is nearly impractical due to constraints such as limited installation space and high costs. \cite{sahin2022evolved} used supervised ML techniques for diagnosis. The authors of this paper applied and compared 13 different ML techniques (Light Gradient Boosting Machine (Light GBM), Random Forest (RF), Gradient Boosting Classifier (GBC), Extra Trees (ET), Quadratic Discriminant Analysis (QDA), Decision Tree (DT), K-Nearest Neighbors (KNN), Ridge Regression (Ridge), Linear Discriminant Analysis (LDA), SVM, Logistic Regression (LR), AdaBoost (ADA), Naive Bayes (NB)) ensembled with Bagging method or with the Blending method according to the accuracy of diagnosis results and execution time of diagnosis. 
While the bagging method is used to reduce the variance of a learning model and combines homogeneous learners. Blending combines heterogeneous learners to produce a stronger model with less biased error than their components. These methods were applied to diagnose faults related to the cooling and lubrication faults (failure of oil jet and insufficient cooling liquid for cylinder cooling). The results showed that with the Bagging ensemble, the Gradient Boosting Classifier (GBC) showed better performance where 98.08\% is shown but with long model execution, while when the blending ensemble method is applied, most successful results were shown when combining with GBC and RF with an accuracy of 98.43\% and 86 sec construction time. \cite{liang2023unsupervised} used an anomaly Transformer NN (TNN) and residual analysis for faults related to the cooling system. The TNN is an unsupervised deep learning used in an autoencoder manner where there is no need for faulty data to build the model, the faulty data is reconstructed using the trained model. The Transformer Autoencoder (TAE) comprises two identical encoder layers with four attention heads each. Unlike traditional Transformer architectures, an efficient reconstruction process is achieved using a Multi-Layer Perceptron (MLP) in the decoder. This AE application of the Transformer model streamlines complexity by excluding shifted features in decoding, relying solely on learned representations for reconstruction. Then the Sequential Probability Ratio Test (SPRT) is used to make decisions about a hypothesis based on a sequential analysis. The sum of squares of normalized residuals (SSNR) follows a Chi-square distribution with k degrees of freedom which is equal to the number of features, and the threshold is derived by inverse cumulative distribution function, used for the reconstruction error. This method is applied on cooling faults and showed stability of TAE performance but it took time in execution and is only tested on one fault type which may not be efficient on other components fault detection.
\vspace{-0.2cm}
\section{Conclusion}\label{sec:conclusion}
\vspace{-0.4cm}
In conclusion, this survey paper has meticulously defined the key subsystems of a marine diesel engine, and introduced various diagnostic techniques, with a particular focus on data-driven approaches. The analysis of recent research papers employing data-driven diagnostic techniques for specific subsystems of marine diesel engines has showcased significant advancements in efficiency and diagnostic performance. However, the observation that many of these methods are predominantly tested on specific faults underscores a limitation in the broader applicability of these approaches. Regrettably, only a limited number of studies have tackled the diagnostic challenges of entire subsystems within marine diesel engines, primarily due to the insufficient availability of comprehensive datasets for comprehensive model training. While promising strides have been made, future research must address this gap and broaden its scope to encompass entire subsystems, aiming to enhance the overall robustness and applicability of diagnostic methodologies in the realm of marine diesel engines. This critical need emphasizes the potential for further advancements in ensuring the reliability and performance of marine diesel engines through comprehensive data-driven diagnostic techniques. This survey paper is planned to be extended into a journal paper, where fault isolation will be discussed and a more in-depth exploration of other diagnostic methods.
\begin{ack}
\vspace{-0.4cm}
This work is done in the framework of a research and development project: "Transformation Numérique du Transport Maritime (TNTM)". This project is funded by BPIFrance, Banque Publique d'Investissement, France.
\end{ack}

\bibliography{ifacconf} 
\vspace{-0.5cm}
\end{document}